\pdfoutput=1
\documentclass[11pt]{article}

\usepackage[]{EMNLP2023}

\usepackage{times}
\usepackage{latexsym}
\usepackage[T1]{fontenc}
\usepackage[utf8]{inputenc}
\usepackage{microtype}
\usepackage{inconsolata}
\usepackage{algorithm}
\usepackage{algpseudocode}
\usepackage{svg}
\usepackage{multirow}
\usepackage{graphicx}
\newcommand*{\dittoclosing}{\textquotedbl}

\title{An End-to-End System for Reproducibility Assessment of Source Code Repositories via Their Readmes}

\author{
  Eyüp Kaan Akdeniz \and
   Selma Tekir \and 
   Malik Nizar Asad Al Hinnawi\\
  Izmir Institute of Technology\\
  Dept. of Computer Engineering\\
35430 Izmir, Turkey\\
  \texttt{kaanakdenz@gmail.com, selmatekir@iyte.edu.tr, malikhinnawi01@gmail.com}}

\begin{document}
\maketitle
\begin{abstract}
Increased reproducibility of machine learning research has been a driving force for dramatic improvements in learning performances. The scientific community further fosters this effort by including reproducibility ratings in reviewer forms and considering them as a crucial factor for the overall evaluation of papers. Accompanying source code is not sufficient to make a work reproducible. The shared codes should meet the ML reproducibility checklist as well. This work aims to support reproducibility evaluations of papers with source codes. We propose an end-to-end system that operates on the Readme file of the source code repositories. The system checks the compliance of a given Readme to a template proposed by a widely used platform for sharing source codes of research. 
Our system generates scores based on a custom function to combine section scores. We also train a hierarchical transformer model to assign a class label to a given Readme. The experimental results show that the section similarity-based system performs better than the hierarchical transformer. Moreover, it has an advantage regarding explainability since one can directly relate the score to the sections of Readme files.
\end{abstract}

\section{Introduction}


The reproducibility crisis is a phenomenon described in that though there is a substantial growth in the number of research papers, the subset that meets the reproducibility criteria is still below expectations. It is a vital issue because one should reproduce the results to judge the claims made in scientific research thoroughly. \citeposs{laurinavichyute2022} work demonstrated that adopting open data policies enhances the reproducibility of studies. They also found that accompanying papers with source codes promotes reproducibility rates.

The increased sharing of code and data is beneficial but requires compliance with well-defined standards. In academia, various standardization mechanisms, including checklists, data sheets, and reproducibility challenges, have been proposed to meet this requirement \cite{gundersen2023}. However, there is no academy-wide uniformity in the definition, measurement, and treatment of reproducibility; instead, there is a growing diversity of views \cite{belz2021}. In addition, the process of verifying compliance with standards is manual, making the processes lengthy and the results less reliable.

Furthermore, the shared codes have some deficiencies. The \citeposs{vandewalle2019} experiment reveals that code availability is still an issue in those papers, as 71\% of the links belonged to valid addresses. Another obstacle to reproducing the same experiments is related to code quality. The lack of a class or interface, dependency on a specific file, and deprecated methods are among the most common issues \cite{mondal2021}. \citet{diaba-nuhoho2021} state that the fundamental aspect of reproducible code is substantial documentation, including open workflows, registered study protocols, and readable and well-maintained code.

\begin{figure}
    \centering
    \includegraphics[width=\columnwidth]{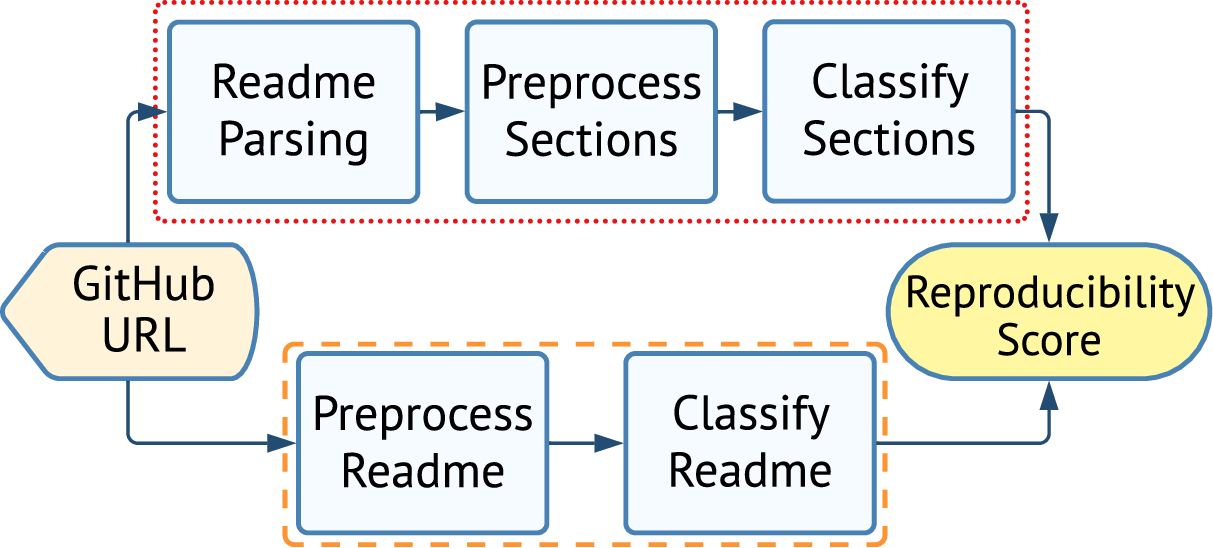}
\caption{End-to-end System Workflow}\label{fig:merged_flow}
\end{figure}

Source codes of machine learning papers are often available in code repositories such as GitHub, Bitbucket, and GitLab. These repositories typically include a "Readme" file in the markdown format, which serves as a directive to guide researchers in reproducing the results reported in the papers. According to \citeposs{obels2019} work, readme files can significantly impact the reproducibility of research as they serve as a bridge between the complex codebase and the research paper, allowing other researchers to understand, replicate, and build upon the reported results. Thus, the quality and comprehensiveness of Readme files are essential.
 
In this study, we introduce a novel, fully-automated end-to-end system to assess the reproducibility of machine learning papers. The system operates by taking a GitHub project link and subsequently generates a reproducibility score based on the content of the project's Readme file. Our workflow (Figure \ref{fig:merged_flow}) includes two alternative models, one that relies on the classification of sections and another based on hierarchical transformers \citep{chalkidis2022}, where the aim is to predict to what extent the associated study is suitable for reproduction by other researchers or practitioners.

Our analysis is based on a readme template proposed by a widely used platform \citep{paperswithcode} for sharing the research code. This template is created by examining existing repositories, identifying those that received the most positive response within the community, and then pinpointing common elements that correlate with popularity \cite{paperswithcode2023}. This template suggests six sections for a readme of a reproducible project: Introduction, Requirements, Pre-trained Models, Training, Evaluation, and Results. It also provides a brief description of each section.

In this work, we propose a comprehensive framework to assess the compliance of \textit{Readme} files with the criteria outlined in the template above. The approach produces a reproducibility score for a given Readme. We perform validation on a hold-out test set of reproducibility-checked NeurIPS2019 papers (\citeposs{paperswithcode2023}). We also provide our system as an open tool\footnote{\url{https://repro-der.streamlit.app/}} with our codes, data, and models\footnote{\url{https://github.com/kaanakdeniz/reproducibility_assessment}}. 

\section{System Workflow}

    Our system is structured around three essential components: Readme parsing, Readme processing, and reproducibility scoring. Figure \ref{fig:merged_flow} illustrates the system workflow. The flow initiates by receiving a GitHub link as input and proceeds through the following stages:

     \subsection{Readme Parsing}
    At this stage, the custom-developed parser tool converts Readme files from the Markdown format into HTML. They are subsequently partitioned into sections, each consisting of a parent header, header, and content. 

    \subsection{Readme Processing}
     The core processing component of the system adopts two distinct methodologies on the parsed and preprocessed sections. The first one performs section classification and generates classification scores for sections. The second method acknowledges the hierarchical structure of the whole Readme to reach a single score for the entire content.

    
    \subsubsection{Section Classification}
        This step automates the measurement of the compliance of a Readme file with the ML reproducibility checklist through the classification of template sections. We utilize a custom scoring function based on section classification to assess template coverage. 
        We use the 'bert-base-uncased' \citep{bertmodel} version of the BERT model \citep{devlin2018} for our section classification model and perform task-specific fine-tuning to improve performance.
        
    
    \subsubsection{Reproducibility Scoring}
    

    Our base formulation for generating reproducibility scores based on section classification is given in Formula \ref{fml:reprod_score1}:
    \begin{equation}\label{fml:reprod_score1}
    \mathrm{R}(C_S, R_c) = \frac{\sum_{S_a \in C_S}^{} \left({\mathrm{max}\left(S_a\right)}\right)}{\mathrm{len}\left(R_c\right)}
    \end{equation}

    Essentially, it's an alignment score between a given Readme ($\mathrm{C_S}$) and the Readme template of the ML reproducibility checklist ($\mathrm{R_c}$). Here, ($\mathrm{S_a}$) represents the array of section classification scores, and we select the maximum ones for every checklist element. A subsequent summation and normalization concerning the checklist length ensure an outcome between $0$ and $1$ to indicate non-reproducible and reproducible, respectively.

    In a second variant of this formulation, section classification is followed by averaging the scores of consecutive sections bearing the same label (see Appendix \ref{app:1} for algorithm details) to exploit the hierarchical structure of Readme. Then, max. taking, summation, and normalization steps are applied.

    \subsubsection{Readme Classification}
    
        
    
        Alternatively, we employ Hierarchical Attention Transformers (HAT) \citep{chalkidis2022} to directly produce a reproducibility score for the entire Readme. The output score is in the range of $0$ to $6$.
    
        The main advantage of this solution is its simplicity - it's an all-in-one process. However, it falls short in terms of explainability. For instance, while the section classification-based system can recognize missing or weak sections as part of its flow, the hierarchical model does not offer such insights.

\section{Data}

 In our experiments, we relied on two datasets: One prepared out of aclanthology.org papers and NeurIPS 2019 top $100$ papers with the maximum star ratings. We used the former primarily for training, while the latter is exclusively reserved for testing the system's performance.

    \subsection{Data Collection-ACL Anthology Papers}
    We collected a total of $47,117$ papers between $2013$ and $2021$ from the aclanthology.org website \citep{aclweb} by a custom-developed crawler. Only $9,300$ of them provided links to their source codes. Out of this subset, we managed to get Readme files for $7,460$ papers that used GitHub for code sharing and whose links were still working.


    After obtaining Readme files, we extracted a structure that includes a header, its parent header (if present), and associated content. We then filtered out sections with irrelevant keywords (see \ref{app:4}) and standardized the text to contain only Latin characters, numbers, and punctuation marks.

    To preserve the structural arrangement of Readme files, we applied different groupings to the header, parent header, and content\footnote{In Readme files, each header may have a distinct level representing a hierarchical structure. Parent-header grouping is merging sections according to their parent headers.}. Overall, we produced $50,472$ individual and $33,193$ parent header-grouped readme sections.

\subsection{Data Labeling}

The ML reproducibility checklist provided by \citet{paperswithcode2023} proposes a Readme template that includes an introduction, requirements, pre-trained models, training, evaluation, and results sections. Accordingly, we determined our multi-class labels for the Readme sections.

We employed automatic labeling techniques to enhance the overall efficiency of the process and due to the superior performances of text similarity-based classification methods' on short texts. Furthermore, we manually annotated a subset of our dataset to ensure label integrity. 




    \subsection{Text Similarity}
    Our text similarity-based labeling makes pairwise comparisons between each readme section and sections within the template. Accordingly, each section is labeled with its most similar pair from the template. To measure the similarity, we generate semantically-rich sentence embeddings \citep{song2020, reimers2019} for sections and use the cosine similarity to relate them. This strategy is well-accepted when the aim is to align with predetermined standards \citep{pineau2020} similar to the baseline template in this work.


    \subsection{Zero-Shot Section Classification}
    Zero-shot learning refers to a paradigm in machine learning in which a model can deliver precise predictions for classes that were not exposed to it during the training phase \citep{xian2019}. Utilizing the "facebook/bart-large-mnli" \citep{zeroshotmodel} pre-trained transformer model based on \citeposs{lewis2020} research, we used zero-shot classification to assign labels to the readme sections. It fits the section classification task well because the classifier will still be capable of identifying sections that do not frequently occur in the training set yet carry semantic relevance \citep{xian2019, rios2018}.

    The results show that the highest level of agreement is achieved with the combination of parent, header, and content section structure for zero-shot and with a header only for text similarity (Appendix \ref{app:6}).

    \subsection{Manual Annotation}
    Furthermore, we manually annotated a subset of the data collection. The subset includes $1050$ data instances with an equal share from each section. To ascertain the level of consistency among the three human annotators, we computed  Weighted Cohen's Kappa \citep{cohen1968} measure. The measure scored $0.59$, which means that humans agreed on the label for $731$ sections.

    \subsection{NeurIPS Papers} \label{sec:neurips}
    This dataset encompasses the top $100$ papers with the maximum star ratings in NeurIPS2019. A \citeposs{paperswithcode2023} report evaluates the papers' repositories with respect to the ML reproducibility checklist. We manually labeled the sections of the Readmes that belong to these papers' repositories. We use this dataset as a hold-out test set for measuring performance.

\section{Results}
We test the effectiveness of our system on the NeurIPS dataset using three different metrics: 1) correlation, 2) agreement, and 3) accuracy. Overall system performance can be represented most effectively through correlation values, while agreement and accuracy rates are more suitable for measuring classification performance. See Appendix \ref{app:5} for runtime details of processes.

    \subsection{Classification-Based System}

    \begin{table}
        \centering
        \scalebox{0.8}{
        \begin{tabular}{|l|c|c|c|} \hline
        \textbf{Type} & \textbf{Corr.} & \textbf{Agr.} & \textbf{Acc.} \\ \hline
        Base & \textbf{0.571} & \textbf{0.547} & \textbf{0.663} \\ \hline
        \begin{tabular}[c]{@{}l@{}}Consecutive\end{tabular} & 0.568 &  \dittoclosing& \dittoclosing \\ \hline
        \end{tabular}
        }
        \caption{System evaluation results based on scoring types.}
        \label{tab:e2e_scoring_types}
    \end{table}
    \begin{table}
        \centering
        \scalebox{0.8}{
        \begin{tabular}{|l|c|c|c|}
        \hline
        \textbf{Method} & \textbf{Corr.} & \textbf{Agr.} & \textbf{Acc.} \\ \hline
        Text Sim. & 0.478 & 0.443 & \textbf{0.643} \\ \hline
        Zero-shot & \textbf{0.486} & \textbf{0.469} & 0.632 \\ \hline
        \end{tabular}
        }
        \caption{System evaluation results based on labeling method of training data.}
        \label{tab:e2e_labeling}
    \end{table} 
    \begin{table}
        \centering
        \scalebox{0.8}{
        \begin{tabular}{|l|c|c|c|} \hline
        \textbf{System} & \textbf{Corr.} & \textbf{Agr.} & \textbf{Acc.} \\ \hline
        Classification & \textbf{0.568} & \textbf{0.528} & \textbf{0.692} \\ \hline
        Hierarchical & 0.495 & 0.404 & 0.300 \\ \hline
        \end{tabular}
        }
        \caption{System evaluation results based on evaluation model.}
        \label{tab:e2e_evalmodel}
    \end{table}

    Table \ref{tab:e2e_scoring_types} presents the section classification-based system's performance using the base formula (Formula \ref{fml:reprod_score1}), and its variant, averaging consecutive section scores. As can be seen, the base formula performs better than the consecutive.
    
    As for comparing labeling techniques, zero-shot outperforms text similarity in correlation and accuracy. However, the text similarity method surpasses zero-shot slightly when evaluated on accuracy (Table \ref{tab:e2e_labeling}).
    
    Upon assessing the average system performance relative to the content of the training data (Appendix \ref{app:7}), we find that the grouped data produces the best results in terms of both correlations (\textit{0.508}) and agreement \textit{(0.493}). As for accuracy, the combination of parent+header+content yields the peak score (\textit{0.667}), slightly higher than that of grouped data.
   
    When we evaluate the effect of the structural content, labeling method, and base vs. consecutive scoring on success rates, we observe that more content is helpful, and grouping the sections boosts the performance. As for the labeling methods, zero-shot yields the top performance (see Appendix \ref{app:3} for the detailed results).

    \subsection{Hierarchical Transformers}    

    In hierarchical transformer training, we utilized the data automatically labeled by text-similarity with the highest human agreement rate (see Appendix \ref{app:6}). The ground-truth labels range from $0$ to $6$, indicating the number of checklist sections in a readme.

    Table \ref{tab:e2e_evalmodel} shows the performances of the section classification-based model and the hierarchical transformer, both trained on identical datasets. The results show that the section classification-based system has superior performance. However, the hierarchical model is still a viable solution because correlation is the primary metric for evaluating the overall performance, and the difference in correlation is the smallest ($15\%.$). It's a holistic approach that assesses the entire readme at once and seems promising for future research.

\section{Conclusion and Future Work}
This work proposes an automated end-to-end system for assessing the readme files of source code repositories with an ML reproducibility checklist template. 

Our proposed framework is an effort to support reproducibility, where academic consensus has yet to be achieved \citep{belz2021}. By utilizing our tool, reviewers and researchers can get feedback about the reproducibility of their projects through readme files.

Future work can enhance reproducibility assessments by automatically evaluating source codes (\citet{trisovic2022}). Furthermore, incorporating automated scientific review \citep{yuan2021} could pave the way toward a comprehensive evaluation of research and its reproducibility.


\section*{Limitations}
Our developed system works with Readme files in English; extension to other languages may require multilingual counterparts for the pre-trained models. Additionally, the system checks the compliance with a specific template. Thus, changes in standards may require the system to be modified and re-validated. Therefore, its functionality might not be suitable for ensuring adherence to varying standards.

Our system assesses the degree to which code repository readmes contribute to reproducibility. It does not evaluate datasets, experimental design, or code provided as part of the repositories. It also does not assert the content quality in the assessed documents.


The automated system provides a layer of assistance, but the critical evaluation and ethical considerations still require human intervention. 

We believe these limitations provide opportunities for further enhancing our system, a precursor for more comprehensive future research reproducibility solutions.

\section*{Acknowledgements}
We gratefully acknowledge Kaggle for their generous GPU resources, HuggingFace for the easy access to pre-trained models, and Papers with Code for their valuable reproducible research guide. 

\bibliography{custom}

\begin{thebibliography}{23}
\expandafter\ifx\csname natexlab\endcsname\relax\def\natexlab#1{#1}\fi

\bibitem[{Belz et~al.(2021)Belz, Agarwal, Shimorina, and Reiter}]{belz2021}
Anya Belz, Shubham Agarwal, Anastasia Shimorina, and Ehud Reiter. 2021.
\newblock \href {https://doi.org/10.18653/v1/2021.eacl-main.29} {A {{Systematic Review}} of {{Reproducibility Research}} in {{Natural Language Processing}}}.
\newblock In \emph{Proceedings of the 16th {{Conference}} of the {{European Chapter}} of the {{Association}} for {{Computational Linguistics}}: {{Main Volume}}}, pages 381--393, {Online}. {Association for Computational Linguistics}.

\bibitem[{Chalkidis et~al.(2022)Chalkidis, Dai, Fergadiotis, Malakasiotis, and Elliott}]{chalkidis2022}
Ilias Chalkidis, Xiang Dai, Manos Fergadiotis, Prodromos Malakasiotis, and Desmond Elliott. 2022.
\newblock \href {https://doi.org/10.48550/arXiv.2210.05529} {An {{Exploration}} of {{Hierarchical Attention Transformers}} for {{Efficient Long Document Classification}}}.

\bibitem[{Cohen(1968)}]{cohen1968}
Jacob Cohen. 1968.
\newblock \href {https://doi.org/10.1037/h0026256} {Weighted kappa: {{Nominal}} scale agreement provision for scaled disagreement or partial credit.}
\newblock \emph{Psychological Bulletin}, 70(4):213--220.

\bibitem[{Devlin et~al.(2018)Devlin, Chang, Lee, and Toutanova}]{devlin2018}
Jacob Devlin, Ming{-}Wei Chang, Kenton Lee, and Kristina Toutanova. 2018.
\newblock \href {http://arxiv.org/abs/1810.04805} {{BERT:} pre-training of deep bidirectional transformers for language understanding}.
\newblock \emph{CoRR}, abs/1810.04805.

\bibitem[{{Diaba-Nuhoho} and {Amponsah-Offeh}(2021)}]{diaba-nuhoho2021}
Patrick {Diaba-Nuhoho} and Michael {Amponsah-Offeh}. 2021.
\newblock \href {https://doi.org/10.1186/s13104-021-05875-3} {Reproducibility and research integrity: The role of scientists and institutions}.
\newblock \emph{BMC Research Notes}, 14(1):451.

\bibitem[{Gundersen et~al.(2023)Gundersen, Coakley, Kirkpatrick, and Gil}]{gundersen2023}
Odd~Erik Gundersen, Kevin Coakley, Christine Kirkpatrick, and Yolanda Gil. 2023.
\newblock \href {http://arxiv.org/abs/2204.07610} {Sources of {{Irreproducibility}} in {{Machine Learning}}: {{A Review}}}.

\bibitem[{{Hugging Face, Inc.}(2019)}]{bertmodel}
{Hugging Face, Inc.} 2019.
\newblock \href {https://huggingface.co/bert-base-uncased} {Bert-base-uncased {$\cdot$} {{Hugging Face}}}.

\bibitem[{{Hugging Face, Inc.}(2020)}]{zeroshotmodel}
{Hugging Face, Inc.} 2020.
\newblock \href {https://huggingface.co/facebook/bart-large-mnli} {facebook/bart-large-mnli}.

\bibitem[{Laurinavichyute et~al.(2022)Laurinavichyute, Yadav, and Vasishth}]{laurinavichyute2022}
Anna Laurinavichyute, Himanshu Yadav, and Shravan Vasishth. 2022.
\newblock \href {https://doi.org/10.1016/j.jml.2022.104332} {Share the code, not just the data: {{A}} case study of the reproducibility of articles published in the {{Journal}} of {{Memory}} and {{Language}} under the open data policy}.
\newblock \emph{Journal of Memory and Language}, 125:104332.

\bibitem[{Lewis et~al.(2020)Lewis, Liu, Goyal, Ghazvininejad, Mohamed, Levy, Stoyanov, and Zettlemoyer}]{lewis2020}
Mike Lewis, Yinhan Liu, Naman Goyal, Marjan Ghazvininejad, Abdelrahman Mohamed, Omer Levy, Veselin Stoyanov, and Luke Zettlemoyer. 2020.
\newblock \href {https://doi.org/10.18653/v1/2020.acl-main.703} {{{BART}}: {{Denoising Sequence-to-Sequence Pre-training}} for {{Natural Language Generation}}, {{Translation}}, and {{Comprehension}}}.
\newblock In \emph{Proceedings of the 58th {{Annual Meeting}} of the {{Association}} for {{Computational Linguistics}}}, pages 7871--7880, {Online}. {Association for Computational Linguistics}.

\bibitem[{Mondal and Roy(2021)}]{mondal2021}
Saikat Mondal and Banani Roy. 2021.
\newblock \href {https://doi.org/10.48550/arXiv.2112.10056} {Reproducibility {{Challenges}} and {{Their Impacts}} on {{Technical Q}}\&{{A Websites}}: {{The Practitioners}}' {{Perspectives}}}.

\bibitem[{Obels et~al.(2019)Obels, Lakens, Coles, Gottfried, and Green}]{obels2019}
Pepijn Obels, Daniel Lakens, Nicholas~A. Coles, Jaroslav Gottfried, and Seth~Ariel Green. 2019.
\newblock \href {https://doi.org/10.31234/osf.io/fk8vh} {Analysis of {{Open Data}} and {{Computational Reproducibility}} in {{Registered Reports}} in {{Psychology}}}.

\bibitem[{{Papers With Code}(2020)}]{paperswithcode2023}
{Papers With Code}. 2020.
\newblock \href {{https://github.com/paperswithcode/releasing-research-code}} {Tips for {{Publishing Research Code}}}.

\bibitem[{{Papers with Code}(2023)}]{paperswithcode}
{Papers with Code}. 2023.
\newblock Papers with code.
\newblock \url{https://paperswithcode.com/}.

\bibitem[{Pineau et~al.(2020)Pineau, {Vincent-Lamarre}, Sinha, Larivi{\`e}re, Beygelzimer, {d'Alch{\'e}-Buc}, Fox, and Larochelle}]{pineau2020}
Joelle Pineau, Philippe {Vincent-Lamarre}, Koustuv Sinha, Vincent Larivi{\`e}re, Alina Beygelzimer, Florence {d'Alch{\'e}-Buc}, Emily Fox, and Hugo Larochelle. 2020.
\newblock \href {https://doi.org/10.48550/arXiv.2003.12206} {Improving {{Reproducibility}} in {{Machine Learning Research}} ({{A Report}} from the {{NeurIPS}} 2019 {{Reproducibility Program}})}.

\bibitem[{Reimers and Gurevych(2019)}]{reimers2019}
Nils Reimers and Iryna Gurevych. 2019.
\newblock \href {https://doi.org/10.18653/v1/D19-1410} {Sentence-{{BERT}}: {{Sentence Embeddings}} using {{Siamese BERT-Networks}}}.
\newblock In \emph{Proceedings of the 2019 {{Conference}} on {{Empirical Methods}} in {{Natural Language Processing}} and the 9th {{International Joint Conference}} on {{Natural Language Processing}} ({{EMNLP-IJCNLP}})}, pages 3980--3990, {Hong Kong, China}. {Association for Computational Linguistics}.

\bibitem[{Rios and Kavuluru(2018)}]{rios2018}
Anthony Rios and Ramakanth Kavuluru. 2018.
\newblock \href {https://doi.org/10.18653/v1/D18-1352} {Few-{{Shot}} and {{Zero-Shot Multi-Label Learning}} for {{Structured Label Spaces}}}.
\newblock In \emph{Proceedings of the 2018 {{Conference}} on {{Empirical Methods}} in {{Natural Language Processing}}}, pages 3132--3142, {Brussels, Belgium}. {Association for Computational Linguistics}.

\bibitem[{Song et~al.(2020)Song, Tan, Qin, Lu, and Liu}]{song2020}
Kaitao Song, Xu~Tan, Tao Qin, Jianfeng Lu, and Tie-Yan Liu. 2020.
\newblock \href {http://arxiv.org/abs/2004.09297} {{{MPNet}}: {{Masked}} and {{Permuted Pre-training}} for {{Language Understanding}}}.

\bibitem[{{The Association for Computational Linguistics}(2023)}]{aclweb}
{The Association for Computational Linguistics}. 2023.
\newblock \href {https://aclanthology.org/} {{{ACL Anthology}}}.

\bibitem[{Trisovic et~al.(2022)Trisovic, Lau, Pasquier, and Crosas}]{trisovic2022}
Ana Trisovic, Matthew~K. Lau, Thomas Pasquier, and Merc{\`e} Crosas. 2022.
\newblock \href {https://doi.org/10.1038/s41597-022-01143-6} {A large-scale study on research code quality and execution}.
\newblock \emph{Scientific Data}, 9(1):60.

\bibitem[{Vandewalle(2019)}]{vandewalle2019}
Patrick Vandewalle. 2019.
\newblock \href {https://kuleuven.limo.libis.be/discovery/fulldisplay?docid=lirias2815281&context=SearchWebhook&vid=32KUL_KUL:Lirias&search_scope=lirias_profile&tab=LIRIAS&adaptor=SearchWebhook&lang=en} {Code availability for image processing papers: A status update}.
\newblock \emph{WIC IEEE SP Symposium on Information Theory and signal Processing in the Benelux}.

\bibitem[{Xian et~al.(2019)Xian, Lampert, Schiele, and Akata}]{xian2019}
Yongqin Xian, Christoph~H. Lampert, Bernt Schiele, and Zeynep Akata. 2019.
\newblock \href {https://doi.org/10.1109/TPAMI.2018.2857768} {Zero-{{Shot Learning}}\textemdash{{A Comprehensive Evaluation}} of the {{Good}}, the {{Bad}} and the {{Ugly}}}.
\newblock \emph{IEEE Transactions on Pattern Analysis and Machine Intelligence}, 41(9):2251--2265.

\bibitem[{Yuan et~al.(2022)Yuan, Liu, and Neubig}]{yuan2021}
Weizhe Yuan, Pengfei Liu, and Graham Neubig. 2022.
\newblock \href {https://doi.org/10.1613/jair.1.12862} {Can we automate scientific reviewing?}
\newblock \emph{J. Artif. Intell. Res.}, 75:171--212.

\end{thebibliography}
\bibliographystyle{acl_natbib}

\appendix

\section{Appendix}
\label{sec:appendix}

    \subsection{Headers of Dropped Sections}\label{app:4}
    By analyzing the frequencies of the section headers, it was determined that the following words were the most frequently found meaningless words in the headers.
    
    \textit{get involved, problems, question, disclaimer, issues, miscellaneous, misc, troubleshoot, reference, references, thoughts, abusive corpus, acknolwedgement, inquiries, changes, ethical guidelines, change logs, citation, cite, credit, contact, licence, acknowledgement, license, referense, contribution, contribute, contributing, author, changelog, faq, citing, news, table of contents, note, links, updates, contributor, todo, acknowledgment, leaderboard, structure, copyright, motivation, acknowledge, what new, bibtex.}
    
    \subsection{Runtime Information}\label{app:5}
    Model training was done on Kaggle with T4x2 GPU.
    Other operations were done on a computer with GTX1650, 16GB, and i7-10750H specifications.
        \subsubsection{Training} 
        Section Classification Model: $\sim$ 1 hour for 3 epoch\\
        Hierarchical Transformers Model: $\sim$ 1.40 hours for 5 epoch 
        \subsubsection{Labeling}
        Zero-shot: $\sim$ 4 hour\\
        Text similarity: $\sim$ 12 min
        \subsubsection{System Evaluation}
        Section Classification Model: $\sim$ 30 sec.\\
        Hierarchical Transformers Model: $\sim$ 45 sec.
        \subsubsection{Readme Parsing}
        Base: $\sim$ 1 sec for 100 readme files.\\
        Grouped: $\sim$3 sec for 100 readme files.

    \subsection{Consecutive Mean Algorithm}\label{app:1}
    \begin{algorithm}[H]
    \begin{algorithmic}
    \Require List of pairs, $L$, where each pair contains a class and a score.
    \Ensure A dictionary, $D$, where each key is a class, and each value is a list of means of consecutive scores of that class.
    \Statex
    \State // Variables:
    \State $R$: A dictionary to store results. Each key is a class, and each value is a list of means of sequential scores of that class.
    \State $P$: The class of the previous pair in the list $L$.
    \State $T$: A temporary list to store the scores of consecutive pairs with the same class.
    \State $(C, S)$: Class and its classification score.
    \Statex
    \State $R \gets \{\}$
    \State $P \gets L[0][0]$
    \State $T \gets []$
    \For{each $(C, S)$ in $L$}
        \If{$C = P$}
            \State Append $S$ to $T$
        \Else
            \State Append $mean(T)$ to $R[P]$ 
            \State $T \gets [S]$
        \EndIf
        \State $P \gets C$
    \EndFor
    \State Append $mean(T)$ to $R[P]$
    \State \Return $R$
    \end{algorithmic}
    \end{algorithm}

   \clearpage 
    \subsection{Evaluation Results Based on Labeling Section Contents}\label{app:7}
    \begin{table}[ht]
    \centering
    \begin{tabular}{|l|r|r|r|} \hline
    \begin{tabular}[c]{@{}l@{}}\textbf{Labeling Section}\\\textbf{Content}\end{tabular} & \textbf{Corr.} & \textbf{Agr.} & \textbf{Acc.} \\ \hline
    Content & 0.485 & 0.448 & 0.635 \\ \hline
    Grouped & \textbf{0.508} & \textbf{0.493} & 0.664 \\ \hline
    Header & 0.404 & 0.401 & 0.600 \\ \hline
    Header + Content & 0.498 & 0.475 & 0.645 \\ \hline
    Parent + Header & 0.503 & 0.471 & 0.614 \\ \hline
    \begin{tabular}[c]{@{}l@{}}Parent + Header \\+ Content\end{tabular} & 0.494 & 0.450 & \textbf{0.667} \\ \hline
    \end{tabular}
    \label{tab:e2e_content}
    \end{table}

    \subsection{Automatic Labeling Performances Based on Section Contents.}\label{app:6}
    \begin{table}[ht]
    \centering
    \begin{tabular}{|l|c|c|c|c|} \hline
    \textbf{} & \multicolumn{2}{c|}{\textbf{Zero Shot}} & \multicolumn{2}{c|}{\textbf{Text Similarity}} \\ \hline
    \textbf{Section Content} & \textbf{Agreement} & \begin{tabular}[c]{@{}c@{}}\textbf{Classification }\\\textbf{Score (Avg.)}\end{tabular} & \textbf{Agreement} & \begin{tabular}[c]{@{}c@{}}\textbf{Similarity }\\\textbf{Score (Avg.)}\end{tabular} \\ \hline
    Header & 0.314 & 0.638 & \textbf{0.345} & 0.275 \\ \hline
    Parent + Header & 0.332 & 0.653 & 0.335 & 0.291 \\ \hline
    Content & 0.236 & 0.661 & 0.227 & 0.368 \\ \hline
    Header + Content & 0.338 & 0.686 & 0.295 & 0.382 \\ \hline
    Parent + Header + Content & \textbf{0.341} & 0.\textbf{700} & 0.300 & \textbf{0.387} \\ \hline
    Grouped & - & 0.678 & - & 0.384 \\ \hline
    \end{tabular}
    \label{tab:agreement}
    \end{table}

    \subsection{Training Results Based on Section Contents}\label{app:2}
    \begin{table}[ht]
    \centering
    \begin{tabular}{|l|c|c|c|c|c|c|} \hline
    \multicolumn{1}{|c|}{} & \multicolumn{3}{c|}{\textbf{Zero Shot}} & \multicolumn{3}{c|}{\textbf{Text Similarity}} \\ \hline
    \textbf{Section Content} & \begin{tabular}[c]{@{}c@{}}Training \\Loss\end{tabular} & \begin{tabular}[c]{@{}c@{}}Validation\\Loss\end{tabular} & Accuracy & \begin{tabular}[c]{@{}c@{}}Training \\Loss\end{tabular} & \begin{tabular}[c]{@{}c@{}}Validation \\Loss\end{tabular} & Accuracy \\ \hline
    Header & 0.60 & 0.83 & 0.71 & 0.54 & 0.74 & 0.72 \\ \hline
    Parent + Header & \textbf{0.52} & \textbf{0.78} & \textbf{0.74} & 0.48 & \textbf{0.65} & \textbf{0.77} \\ \hline
    Content & 0.64 & 0.83 & 0.71 & 0.51 & 0.73 & 0.73 \\ \hline
    Header + Content & 0.61 & 0.85 & 0.70 & 0.50 & 0.75 & 0.72 \\ \hline
    Parent + Header + Content & 0.56 & 0.80 & 0.72 & \textbf{0.47} & 0.71 & 0.74 \\ \hline
    Grouped & 0.62 & \textbf{0.78} & 0.73 & 0.55 & 0.73 & 0.74 \\ \hline
    \end{tabular}
    \end{table}

    \clearpage
    \subsection{Section Classification-Based System's Evaluation Results}\label{app:3}
\begin{table}[ht!]
\centering
\scalebox{0.95}{
\begin{tabular}{|c|c|c|c|c|c|} \hline
\begin{tabular}[c]{@{}c@{}}\textbf{Labeling}\\\textbf{Method}\end{tabular} & \textbf{Labeling Content} & \begin{tabular}[c]{@{}c@{}}\textbf{Scoring Type}\end{tabular} & \textbf{Correlation} & \textbf{Agreement} & \textbf{Accuracy} \\ \hline
\multirow{12}{*}{Text Sim.} & \multirow{2}{*}{Content} & Base & 0.549 & 0.521 & 0.665 \\ \cline{3-6}
 &  & Consecutive & 0.554 & 0.521 & 0.665 \\ \cline{2-6}
 & \multirow{2}{*}{Grouped} & Base & 0.579 & 0.542 & 0.697 \\ \cline{3-6}
 &  & Consecutive & 0.581 & 0.542 & 0.697 \\ \cline{2-6}
 & \multirow{2}{*}{Header + Content} & Base & 0.578 & 0.523 & 0.685 \\ \cline{3-6}
 &  & Consecutive & 0.571 & 0.523 & 0.685 \\ \cline{2-6}
 & \multirow{2}{*}{Parent + Header + Content} & Base & 0.568 & 0.528 & 0.692 \\ \cline{3-6}
 &  & Consecutive & 0.569 & 0.528 & 0.692 \\ \cline{2-6}
 & \multirow{2}{*}{Parent + Header} & Base & 0.602 & 0.613 & 0.668 \\ \cline{3-6}
 &  & Consecutive & 0.597 & 0.613 & 0.668 \\ \cline{2-6}
 & \multirow{2}{*}{Header} & Base & 0.497 & 0.479 & 0.637 \\ \cline{3-6}
 &  & ~Consecutive & 0.473 & 0.479 & 0.637 \\ \hline
\multirow{12}{*}{Zero-Shot} & \multirow{2}{*}{Content} & Base & 0.582 & 0.563 & 0.662 \\ \cline{3-6}
 &  & Consecutive & 0.586 & 0.563 & 0.662 \\ \cline{2-6}
 & \multirow{2}{*}{Grouped} & Base & 0.651 & \textbf{0.648} & 0.697 \\ \cline{3-6}
 &  & Consecutive & \textbf{0.661} & \textbf{0.648} & 0.697 \\ \cline{2-6}
 & \multirow{2}{*}{Header + Content} & Base & 0.631 & 0.631 & 0.665 \\ \cline{3-6}
 &  & Consecutive & 0.626 & 0.631 & 0.665 \\ \cline{2-6}
 & \multirow{2}{*}{Parent + Header + Content} & Base & 0.617 & 0.556 & \textbf{0.698} \\ \cline{3-6}
 &  & Consecutive & 0.624 & 0.556 & \textbf{0.698} \\ \cline{2-6}
 & \multirow{2}{*}{Parent + Header} & Base & 0.594 & 0.540 & 0.608 \\ \cline{3-6}
 &  & Consecutive & 0.587 & 0.540 & 0.608 \\ \cline{2-6}
 & \multirow{2}{*}{Header} & Base & 0.399 & 0.419 & 0.587 \\ \cline{3-6}
 &  & Consecutive & 0.383 & 0.419 & 0.587 \\ \hline
\end{tabular}
}
\end{table}

    We evaluated a factorial design of modeling choices: Labeling method, data content, and scoring type, ending in $24$ in different ways. The results show that the system with the consecutive scoring using the classification model trained on grouped sections that are labeled with the zero-shot method gives the highest correlation (\textit{0.661}) and agreement (\textit{0.648}) value. This combination is still good performing for accuracy, where the best-performing choice becomes parent+header+content input instead of grouped sections. On the other hand, the least successful system in terms of correlation and accuracy rates was the one with consecutive scoring using the classification model trained on the section headers labeled by zero-shot methods. The combination of text similarity and header+content gives the lowest agreement score.
\end{document}